\documentclass[letterpaper, 10 pt, journal, twoside]{IEEEtran}
\usepackage{enumitem}
\usepackage{balance}
\usepackage[font={small}]{caption}
\usepackage{subcaption}
\usepackage{array}
\usepackage{textcomp}
\usepackage{mathtools, nccmath}
\usepackage{graphicx}
\usepackage{amsfonts}
\usepackage{amsmath, amssymb, amsthm}
\usepackage{algorithm}
\usepackage{algorithmic}
\usepackage{hyperref}
\usepackage{tikz}
\usepackage{float}
\usepackage{arydshln}
\usepackage{multirow}
\usepackage{bm}
\usepackage{epstopdf}
\usepackage{cite}
\usepackage{siunitx}
\usepackage{esvect}
\usepackage[normalem]{ulem}
\ifCLASSINFOpdf
\else
\fi

\newtheoremstyle{mystyle}
  {}
  {}
  {}
  {}
  {\bfseries}
  {.}
  { }
  {\thmname{#1}\thmnumber{ #2}\thmnote{ (#3)}}
\theoremstyle{mystyle}
\newtheorem{remark}{Remark}

\usepackage{amsthm}

\title{Collaborative Navigation and Manipulation of a Cable-towed Load by Multiple Quadrupedal Robots} 
\author{Chenyu Yang$^{*,1}$, Guo Ning Sue$^{*,1}$, Zhongyu Li$^{*,1}$, Lizhi Yang$^{1}$, Haotian Shen$^{1}$, Yufeng Chi$^{1}$, \\ Akshara Rai$^{2}$, Jun Zeng$^{1}$, Koushil Sreenath$^{1}$
\thanks{$^*$ Authors contributed equally.} 
\thanks{$^1$ Department of Mechanical Engineering, University of California, Berkeley. yangcyself@gmail.com, \{gusue418, zhongyu\_li, lzyang, dalyshen, chiyufeng, zengjunsjtu, koushils\}@berkeley.edu}
\thanks{$^2$ Meta AI Research, akshararai@fb.com}
}

\begin{document}
\maketitle

\begin{abstract}
This paper tackles the problem of robots collaboratively towing a load with cables to a specified goal location while avoiding collisions in real time. 
The introduction of cables (as opposed to rigid links) enables the robotic team to travel through narrow spaces by changing its intrinsic dimensions through slack/taut switches of the cable.
However, this is a challenging problem because of the hybrid mode switches and the dynamical coupling among multiple robots and the load. Previous attempts at addressing such a problem were performed offline and do not consider avoiding obstacles online. 
In this paper, we introduce a cascaded planning scheme with a parallelized centralized trajectory optimization that deals with hybrid mode switches. We additionally develop a set of decentralized planners per robot, which enables our approach to solve the problem of collaborative load manipulation online.
We develop and demonstrate one of the first collaborative autonomy framework that is able to move a cable-towed load, which is too heavy to move by a single robot, through narrow spaces with real-time feedback and reactive planning in experiments.
\end{abstract}

\section{Introduction}
\begin{figure}[t]
    \centering
    \includegraphics[width=0.95\linewidth]{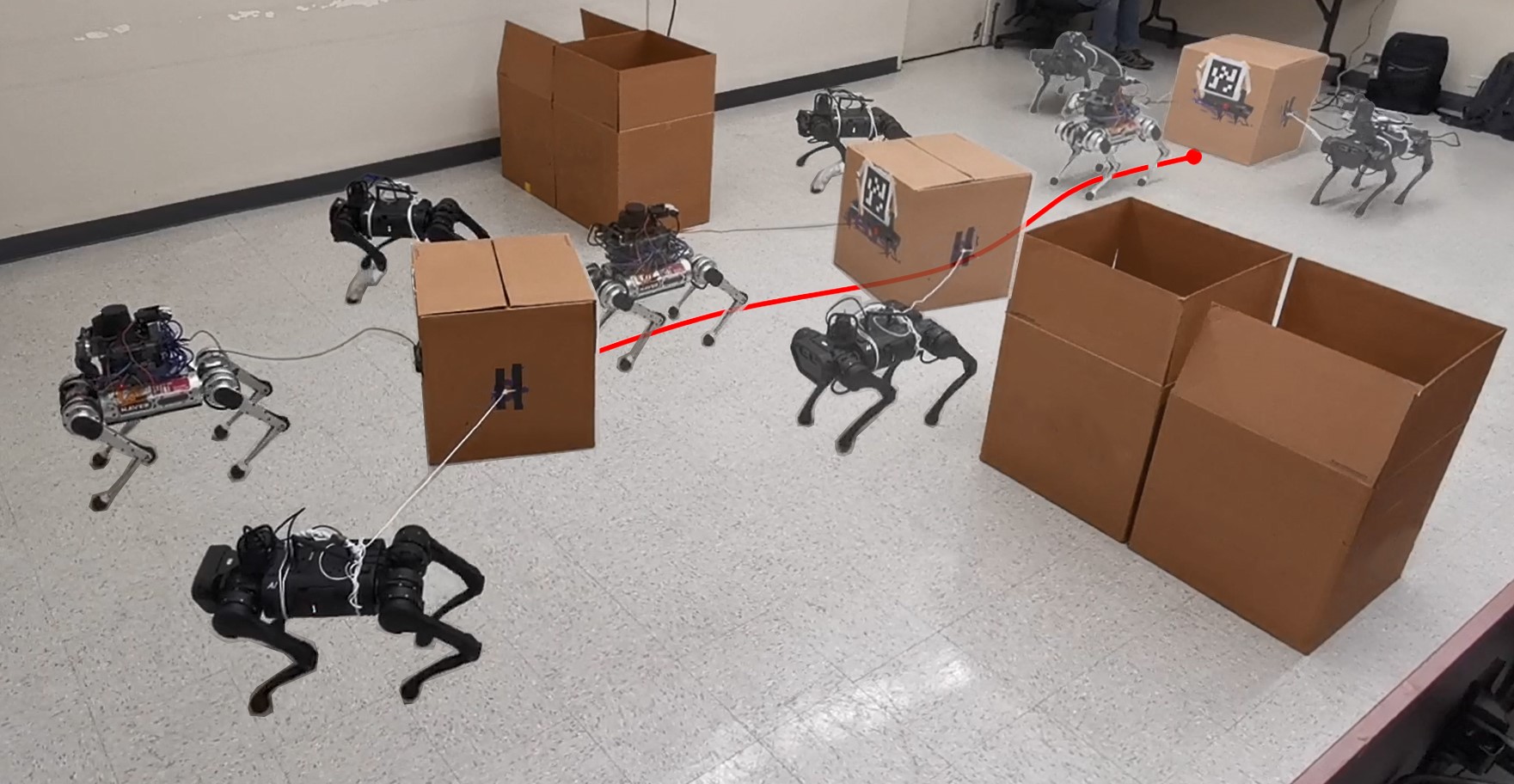}
    \caption{A superimposed image of snapshots of a cable-towed load that is manipulated by a team of three quadrupedal robots traveling in a cluttered environment. Lighter snapshots are earlier in time. The proposed real-time autonomy framework enables a collaborative manipulation task with multiple quadrupedal robots to navigate around the obstacles through hybrid mode switches depending on the cables being taut or slack.}
    \label{fig:main_figure}
    \vspace{-0.3cm}
\end{figure}

Quadrupedal robots have demonstrated agile maneuverability through their versatile locomotion skills. The quadrupedal robot community is active with the recent exciting progress in hardware design, control, and planning~\cite{katz2019mini, kim2019highly, yang2020dynamic, gilroy2021autonomous}.
However, most of the existing research on quadrupeds focuses on a single robot, with just a few approaches that study collaboration between multiple quadrupeds.
Collaboration between multiple quadrupedal robots can be useful in cases where large number of \textit{small} general purpose quadrupeds are available, rather than \textit{larger} special purpose robots.  
The robots can team together to achieve collaborative tasks that can not be achieved by individual robots.
This could be useful in last mile delivery, rapid disaster response, or even extraterrestrial operations.
There is a large body of literature tackling multi-agent autonomy with policy interaction where the behavior of one agent will not physically affect the dynamics of the others~\cite{kanazawa2019adaptive}.
The physical interaction among multiple agents, however, would be advantageous since robots can physically contribute jointly to one task, such as moving a load.
Moreover, the introduction of a cable instead of a rigid link allows the system to travel through narrow spaces by letting the cable switch between slack and taut modes~\cite{xiao2021robotic}.
However, control and planning problems on the collaboration tasks with such hybrid physical interaction among multiple robots are very challenging. This is due to the hybrid mode switches arising from physical interactions and the coupled multi-agent dynamics that need to be considered.
In this paper, we seek to learn the viability of using multiple quadrupedal robots to collaborate and pull a load via cables to travel through cluttered environments in real-time, as demonstrated in Fig.~\ref{fig:main_figure}.

\subsection{Related Work}
\begin{figure*}[!htp]
    \centering
    \includegraphics[width=0.9\linewidth]{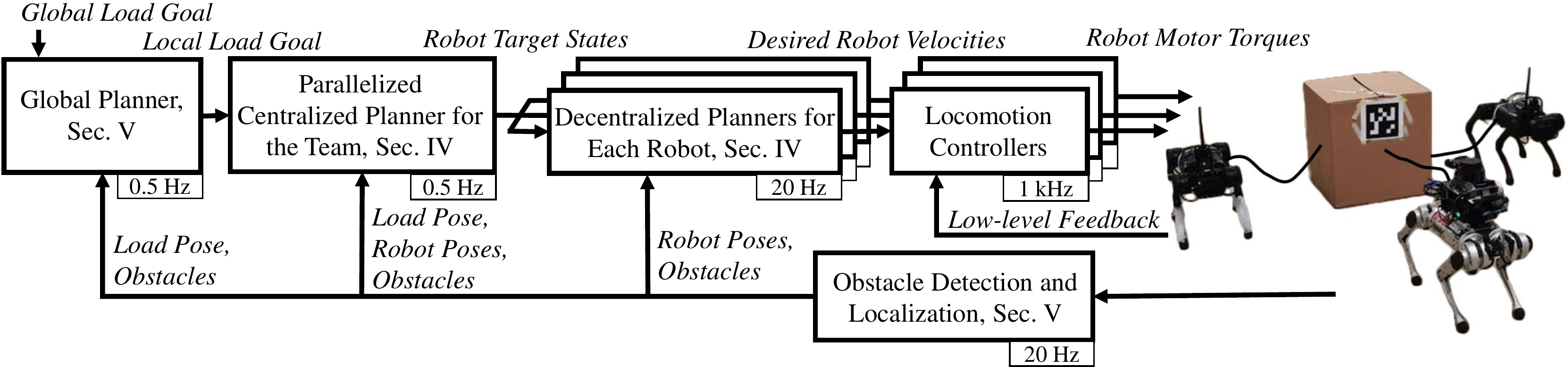}
    \caption{The proposed collaboration autonomy of multiple quadrupedal robots using cables to move a load through cluttered environments. After a global goal for the load is specified, an A* based global planner firstly finds a collision-free path from the current position of the load. It is followed by a centralized planner that solves for multiple trajectories in parallel, each with different hybrid modes, to obtain a hybrid trajectory for the entire robot team and the load.
    Later, each robot runs an optimization-based planner to follow the planned trajectory in a decentralized manner, and has a locomotion controller that tracks the velocity commands from the decentralized planner and maintains gait stability. Moreover, each robot leverages a 2D LiDAR to perceive the environment and detect obstacles.
    A single robot carries an RGB-Depth camera on its back to keep track of the pose of the towed-load.}
    \label{fig:frame_work}
    \vspace{-0.4cm}
\end{figure*}

\subsubsection{Collaborative Tasks with Hybrid Physical Interaction}
Interactive tasks between robots have been popular in recent years.
These works usually involve the study of decision making for control and planning among multiple players (often called agents)~\cite{liniger2019noncooperative, cleac2021algames}. 
One scenario of the multi-agent task is collaborative interaction.  
Physical interactions among agents could be useful in some scenarios where each agent can contribute jointly for a single task.
In these problems, the interactive forces among these agents are important but typically hard to model. Physical interactions among agents can be classified into three types: rigid, soft and hybrid interactions. They are applied to manipulation tasks for robotic arms~\cite{tung2021learning, sanchez2021four}, aerial robots~\cite{wang2018cooperative, sanalitro2020full} and mobile robots~\cite{cheng2009cooperative, wang2016kinematic}.
Rigid interactions can be found commonly in robotic arm manipulation~\cite{losey2020learning, tung2021learning, sanchez2021four} or through rigid linkages~\cite{wang2016kinematic, loianno2017cooperative, culbertson2021decentralized}, which
makes maneuverability in narrow spaces challenging.
Soft interactions on the other hand introduce agility for manipulation tasks with a soft gripper~\cite{wang2017prestressed} or deformable objects~\cite{herguedas2019survey}, but the system dynamics are complex.
Finally, \textit{hybrid physical interactions} are usually achieved with a cable-towed mechanism for moving a load~\cite{kim2018bound, sanalitro2020full, cardona2021non}, which can fully exploit the capabilities of all collaborative agents.
This cable-towed mechanism provides geometry-constrained hybrid interactions as the distance between two attached objects is always equal to or smaller than the cable length.
However, control and planning become much more challenging due to highly nonlinear dynamics. Furthermore, hybrid modes need to be considered when the cables switch between taut and slack.
Due to these challenges, existing work generally focuses on non-slack cable manipulation tasks~\cite{sanalitro2020full, cardona2021non}.
Notice that manipulation mechanisms using variable-length cables are also applicable~\cite{bhattacharya2015topological, zeng2019geometric} in a narrow space, but it would introduce additional degrees-of-freedoms (DoFs) making the control problem more difficult.

\subsubsection{Control \& Planning for Hybrid Manipulation}
Recently, hybrid manipulation for a cable-suspended system has been analyzed theoretically and experimentally.
The control problems of hybrid manipulation have been analyzed through different aspects, such as control equilibrium~\cite{cheng2009cooperative, tognon2018aerial}, flatness~\cite{lee2013geometric, wu2014geometric} and controllability~\cite{Kotaru2020multiple}.
The adaptive control problem for cable-based manipulation tasks with a single quadrupedal robot has also been discussed in~\cite{minniti2021adaptive}.
In the path planning problem for hybrid manipulation, mixed-integer optimization~\cite{tang2015mixed} or general nonlinear programming with force-based complementarity constraints~\cite{zeng2020differential} can be applied but these approaches suffer from nonlinear hybrid dynamics and can only be solved offline.
Recently, authors in~\cite{xiao2021robotic} applied a mixed-integer programming in the local planner which is deployed in real-time with a single robot manipulation task via a cable.
However, all of the previous approaches for solving optimization problems for hybrid mode switching cannot be applied directly to multiple robots for collaborative tasks due to the complexity of a multi-agent system even with simplified dynamics.
Our proposed autonomy framework solves the collaborative hybrid manipulation problem in real-time through cascaded planners, \textit{i.e.}, a parallelized centralized planner followed by a set of decentralized planners.

\subsection{Contributions}
The main contribution of this paper is the development of a real-time collaborative autonomy to manipulate a cable-towed load. 
This autonomy utilizes multiple robots and hybrid physical interaction to manipulate and navigate a cable-towed load through cluttered spaces. 
This is a challenging problem due to the high-dimensional nonlinear dynamics of each robot along with the multiple hybrid modes introduced by the cables. 
In order to address this, we introduce a parallelization framework for optimization to deal with hybrid mode switches, which makes solving trajectory optimization for multiple robots with hybrid dynamics computationally feasible online. With current computational hardware, our approach can scale up to $6$ robots for online operations and for larger number of robots for offline computation.
Such a parallelized optimization solves for different trajectories with different hybrid modes simultaneously, and an optimal trajectory is then selected for the centralized plan for the robot team and the load.  
Our proposed cascaded planning framework leverages global planning, centralized planning, decentralized planning and real-time sensor feedback.
We demonstrate this working experimentally on multiple quadrupedal robots that collaborate together to manipulate a cable-towed load, which is so heavy that a single robot cannot move it, with hybrid mode switches in order to navigate the load to the desired location while avoiding obstacles in real-time. 
This paper serves as a step forward towards reliable collaborative autonomy using robot teams to manipulate and transport cable-towed heavy load in real life.

\label{sec:Introduction}

\section{Overview of Collaborative Autonomy}
\label{sec:framework}

In this section, we present an overview of the entire framework shown in Fig.~\ref{fig:frame_work} that leverages multiple quadrupedal robots to use cables to move a load to a goal location while collaborating with each other and avoiding obstacles.
Besides the controller for each quadruped, the collaboration autonomy for the load carrying task comprises of two main parts: planning and perception for both robots and the load.  

A collision-free path to the given goal location for the cable-towed load is first generated by a global planner based on A* search as introduced in Sec.~\ref{sec:system}.
Later, a local goal position for the towed-load is located on the global path which updates during each replanning of the centralized trajectory planner.
As will be discussed in Sec.~\ref{subsec:centralized_planner}, the centralized planner computes the trajectories for the robots and the load while considering the dynamics of the entire system to avoid obstacles en-route to the given goal position.
Moreover, the hybrid mode switching of the cables during the load manipulation is also determined by this planner.
The cable of each robot in the team can be either slack or taut, and the optimization involving such hybrid modes and coupled dynamics is computationally challenging to solve online due to the increasing number of robots.  
Therefore, in this autonomy, we leverage a parallelized trajectory optimization, \textit{i.e.}, synchronously running multiple optimizations with different combinations of hybrid modes of the cables, and thereby achieve a replanning rate of 0.5~Hz. 
Details will be discussed in Sec.~\ref{subsec:parallel_opt}.
For each robot in the team, a decentralized planner is developed in Sec.~\ref{sec:decentralized_planner} to track the desired trajectory generated from the centralized planner while avoiding collisions with the obstacles, other robots, and the load.
This decentralized planner updates at 20~Hz and outputs the robot's desired sagittal and lateral walking speeds with turning yaw rate for the locomotion controller to generate motor torques.

For perception, each of the robot in the team is equipped with a 2D LiDAR. 
One robot also has a tracking camera on its front and carries a RGB-Depth camera attached to a 2 Degree-of-Freedom~(DoF) gimbal on its back to detect and follow the load.
Each robot is able to estimate its own odometry at 20~Hz. 
Additionally, the 2D LiDAR is also leveraged to detect the surrounding obstacles at 5~Hz. Such system integration is introduced in Sec.~\ref{sec:system}.

In this way, with real-time feedback, the collaboration autonomy is able to achieve the task of towing a load to global goal while traveling through cluttered spaces by reactively avoiding obstacles with hybrid mode switching of cables.
The proposed framework is scalable to more general mobile robot teams where each robot can move omni-directionally, but we focus on quadrupeds that are challenging but more agile.

\section{Dynamic Model of Collaborative Quadrupeds with Cable-towed Load}
\label{sec:model}
In this section, we will first develop the dynamic model of the multiple quadrupedal robots with a cable-towed load. This model will be used by the centralized planner.

\subsection{Configuration of the Collaborative Team with Load}

\begin{figure}[t]
    \centering
    \includegraphics[width=0.65\linewidth]{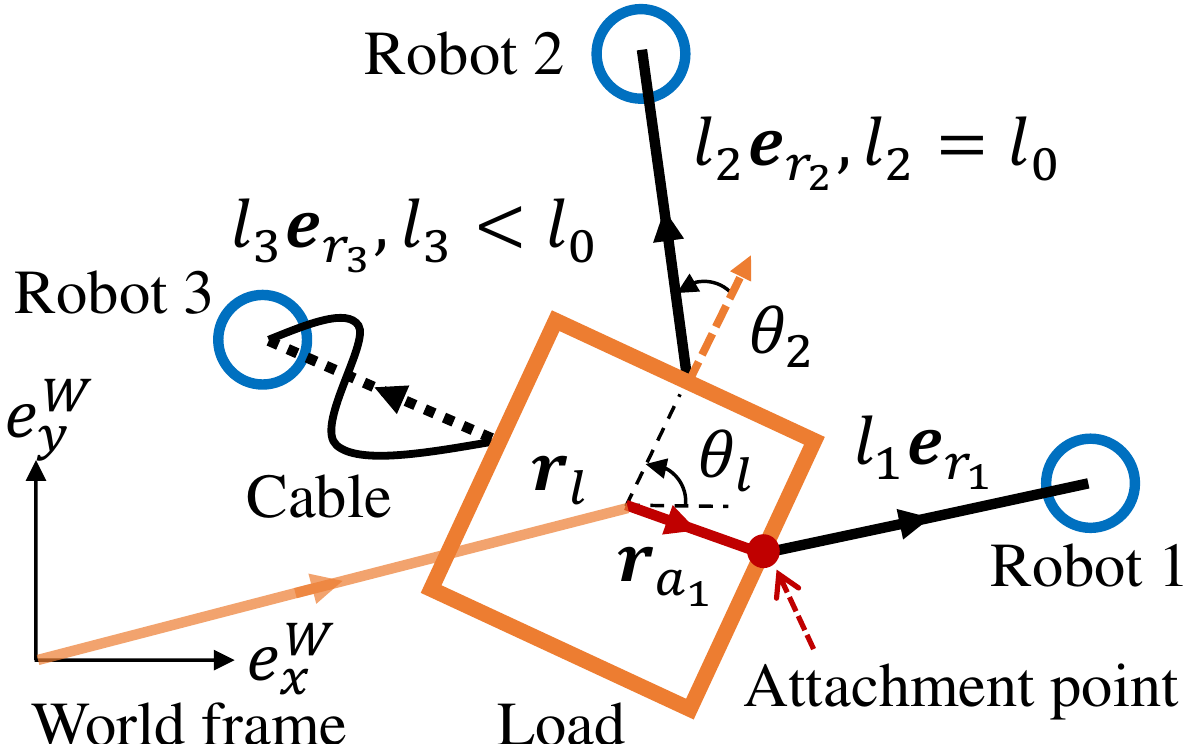}
    \caption{Configuration for multiple robots with a cable-towed load. The load is modeled as a rigid body. Each robot, simplified as a point mass, tows the load with a cable attached to a surface of the load. The cable is taut when the robot-load separation $l$ is equal to cable length $l_0$ while being slack when the separation is less than $l_0$.}
    \label{fig:box_dog_config}
    \vspace{-0.3cm}
\end{figure}

As illustrated in Fig~\ref{fig:box_dog_config}, the system consists of a load and $n$ robots connected to the load via $n$ cables of length $l_0$. 
The system can be simplified and modeled by regarding the load as a rigid body, and robots as point-masses respectively. 
Specifically, the system has the following configuration space:
{\small
\begin{align}
\mathbf{q} &\coloneqq [\mathbf{q}^T_l, \theta_1, l_1 \dots \theta_n, l_{n}]^T \in \mathbb{R}^{3+2n} \label{equ:defn-q}\\
\mathbf{x} &\coloneqq [\mathbf{q}^T, \dot{\mathbf{q}}^T]^T \in \mathbb{R}^{6+4n}\label{equ:defn-x} \\
\mathbf{x}_{r_i} &\coloneqq [x_{r_i}, y_{r_i}, \phi_{r_i}, \dot{x}_{r_i}, \dot{y}_{r_i}, \dot{\phi}_{r_i}]^T \in \mathbb{R}^6, \ \forall i=1,\dots,n \label{equ:defn_robot_x} 
\end{align}}where $\mathbf{q}_{l} = [x_{l}, y_{l}, \theta_{l}]^T \in SE(2)$ which represents the load configuration in the world frame, $\mathbf{x}$ represents the system state, and $\mathbf{x}_{r_i}$ denotes as the $i$-th robot state. Moreover,  $\mathbf{r}_l = [x_l, y_l]^T \in \mathbb{R}^2$ is denoted as the position vector of the load.
The $i$-th cable is connected between the $i$-th attachment point on the load surface and the $i$-th robot. According to Fig.~\ref{fig:box_dog_config}, the vector $\mathbf{r}_{a_i}$ goes from the load center to the attachment point of the $i$-th cable.  
The robot position, as a point mass, can represented by its associated cable orientation $\theta_i$ and distance $l_i$ with respect to the attached surface of the load.  
Furthermore, we introduce an unit vector $\mathbf{e}_{r_i}=[\cos(\theta_l+\theta_i),\sin(\theta_l+\theta_i)]^T \in S^1$ to denote the direction from the load surface to the robot. 
In this way, the $i$-th robot position vector $\mathbf{r}_{r_i}=[x_{r_i}, y_{r_i}]^T$ in the world frame can be defined through forward kinematics.

Additionally, we make two assumptions in the system model to simplify the problem. First, the cables are massless and can not be stretched. Second, each cable is attached at the geometric center of the corresponding robot base. 
Therefore, $l_i \leq l_0, \forall i=1 \dots n$ is a constraint for a valid configuration. 
As shown in Fig.~\ref{fig:box_dog_config}, when $l_i < l_0$, the $i$-th cable is in slack mode, otherwise, the cable is taut.
Moreover, we further assume that the mass of the load is uniformly distributed, \textit{i.e.}, the load's Center-of-Mass~(CoM) is its geometric center.
The robot shape will be considered by the decentralized planners. 

\subsection{Hybrid Dynamics Model}
\label{subsec:hybrid-dynamic-model}

We next define the tension of the $i$-th cable as $T_i \in \mathbb{R}$ and $T_i \geq 0$, and a vector $\mathbf{t}=[T_1, T_2, \dots, T_n]^T$ for tensions on $n$ cables.
Therefore, the system dynamics $\dot{\mathbf{x}} = f(\mathbf{x}, \mathbf{u},\mathbf{t})$ that consists of $n$ robot and a cable-towed load can be written as
{\small
\begin{align}
    m_l \ddot{\mathbf{r}}_l &= \sum_{i=1}^{n} T_i \mathbf{e}_{r_i} - \mu_{tan} m_l g \frac{\dot{\mathbf{r}}_l}{||\dot{\mathbf{r}}_l||_2} \label{eq:model-dyn-pL}\\
    I_l \ddot{\theta}_l &= \sum_{i=1}^{n} \mathbf{r}_{a_i} \times T_i\mathbf{e}_{r_i} - \mu_{tor} m_l g \operatorname{sgn}(\dot{\theta_l})\label{eq:model-dyn-th}\\
    m_{r_i} \ddot{\mathbf{r}}_{r_i} &= \mathbf{u}_{i} - T_i \mathbf{e}_{r_i} \label{eq:model-dyn-p_ri},
\end{align}}where $m_l$, $I_l$ are the mass and inertia of the load, $\mu_{tan}$, $\mu_{tor}$ are the tangential and torsional friction coefficients of the contact between the load and ground, respectively. 
Moreover, $m_{r_i}$ is the mass of the $i$-th robot while $\mathbf{u}_i$ is the virtual input contributing to the robot acceleration.

The system with cables is hybrid by its nature. 
The tension in a cable exists when the cable is taut, \textit{i.e.} $l_i=l_0$. 
On the contrary, if the cable is slack, \textit{i.e.}, $l_i < l_0$, the tension in the cable will be zero $T_i=0$. 
The relationship between cable tensions $T_i$ and distances $l_i$ between robot and the load can be characterized by a complementarity constraint:
{\small
\begin{equation}
    \label{eq:model:complementary_cons}
    T_i(l_0-l_i)=0, \text{and}~T_i\geq0,~l_0-l_i\geq0,
\end{equation}}which ensures the cable tension and the change of cable length cannot be non-zero at the same time. In this way, the hybrid dynamics model of the system with multiple robots and a cable-towed load can be explicitly obtained. 
This model is then used in the optimization-based planners in the next section.

\section{Optimization-based Trajectory Planners}
\begin{figure}
    \centering
    \includegraphics[width=0.7\linewidth]{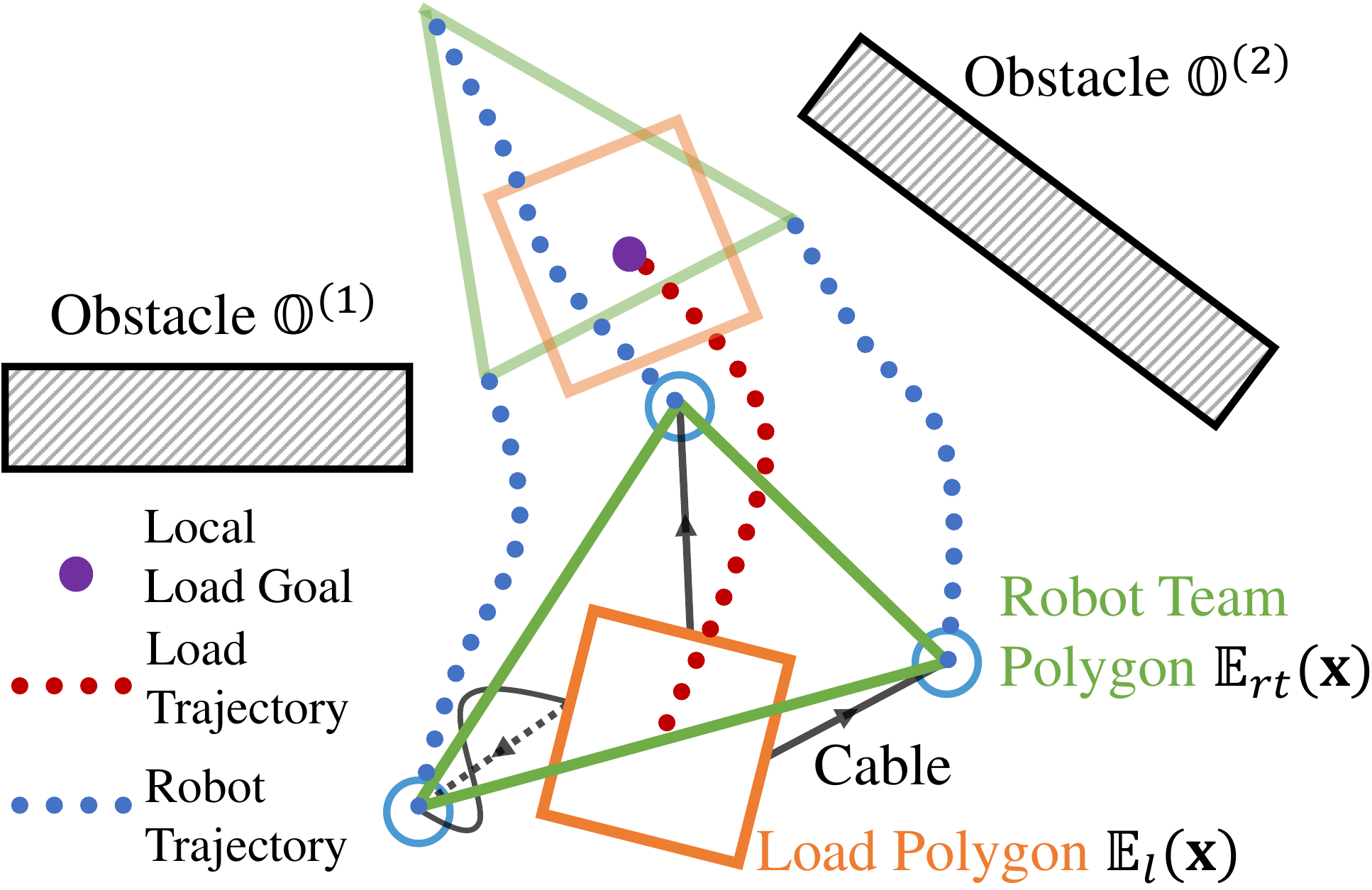}
    \caption{Illustration for trajectory optimization of the centralized planner. Once the local goal is obtained from the global path, marked as purple, the centralized planner will generate collision-free trajectories for the robots and the load. The obstacle avoidance problem can be considered as avoiding collision between the convex polygon of the obstacle and the union of the convex polygon that covers the robot team and load convex polygon.}
    \label{fig:centeralized_planner}
    \vspace{-0.3cm}
\end{figure}

Using the model developed in the previous section, we now develop an optimization-based centralized trajectory generator that considers the above system dynamics with hybrid mode switches to plan for trajectories for robots and the towed load.
In order to solve the optimization online, we further introduce a method to parallelize the optimization for hybrid dynamics.
This parallelized centralized planner serves as the first layer of planning algorithm. We also introduce the decentralized reactive planner in this section.

\subsection{Convex Representation for the System and Obstacles}~\label{subsec:convex_polygon}
In order to tackle the optimization-based collision avoidance problem, we utilize convex polygons to represent the obstacles and the collaborative autonomy that includes the robot team and the cable-towed load. 
We consider collision avoidance as a non-intersection with a minimum distance between two convex polygons~\cite{zhang2020optimization}.

For the collaborative autonomy that has three robots in the team with a cable-towed rectangular load, we choose two convex polygons to cover the system.
One is the triangle formed by connecting the robot team, $\mathbb{E}_{rt}$, and the other is the rectangle that is the shape of the load, denoted by $\mathbb{E}_l$.
An example of those two convex polygons is illustrated in Fig.~\ref{fig:centeralized_planner}.
In this way, the system in any configuration can be represented by the union of the convex robot team polygon and convex load polygon, which can be written as $\mathbb{E}(\mathbf{x}) = \mathbb{E}_{rt}(\mathbf{x}) \cup \mathbb{E}_l(\mathbf{x})$,
{\small
\begin{equation}
\label{eq:ego_polygon}
 \mathbb{E}(\mathbf{x}){=}\{y\in \mathbb{R}^2~|~G_{rt}(\mathbf{x})y{\leq}g_{rt}(\mathbf{x}) \ \operatorname{or} \ G_l(\mathbf{x})y{\leq}g_l(\mathbf{x})\},
\end{equation}}where $G_{rt}\in \mathbb{R}^{n_{rt}\times 2}$, $g_{rt}\in \mathbb{R}^{n_{rt}\times 1}$, $G_l\in \mathbb{R}^{n_l\times 2}$ and $g_l\in \mathbb{R}^{n_l\times 1}$ with $n_{rt}$ being the number of edges in the robot team polygon, and $n_l$ being the number of edges in the load polygon.
In this paper, $n_{rt}=3$ and $n_l=4$. 
Similarly, the $m$-th obstacle can be represented by a compact convex set $\mathbb{O}^{(m)}=\{y\in \mathbb{R}^2~|~A^{(m)}y~\leq b^{(m)}\}$, as shown in Fig.~\ref{fig:centeralized_planner}. 
\begin{remark}\label{remark:triangulation}
The robot team polygon cannot be guaranteed to be convex when it consists of more than three agents. In this case, we can either introduce a triangulation of this polygon or impose a convex constraint during trajectory optimization as shown in~\cite{zeng2020differential}.
\end{remark}

\subsection{Parallelized Optimization for Hybrid Mode Switches}~\label{subsec:parallel_opt}
As discussed in Sec.~\ref{subsec:hybrid-dynamic-model}, there exists hybrid mode switching in the cables during the collaborative manipulation task, which is often described by a complementarity constraint between the cable length and cable tension via~\eqref{eq:model:complementary_cons}.
Such constraints in trajectory optimization often makes the problem not scalable and hard to be solved online, especially as the number of robots increases.
Therefore, inspired by~\cite{he2022autonomous}, we introduce a parallelized  optimization scheme where we simultaneously run multiple optimizations each with a different combination of the hybrid modes. 
The potential hybrid mode combination, \textit{i.e.}, binary vector $\bm{\delta} \in \mathbb{Z}^n$, can be selected from $\{0,1\}^{n}$ where $0$ represents the slack mode, $1$ represents the taut mode, and $n$ is the number of agents.
After obtaining the $2^n$ trajectories solved in parallel, we can select the best among them based on selected metrics such as the objective value, distance to obstacles, etc.
\begin{remark}
Although we fix the hybrid mode during one single trajectory optimization, the hybrid mode may change between two adjacent replanned trajectories based on the different choices of hybrid mode for the optimal trajectory.  
Therefore, the system is still able to utilize the hybrid mode switches at a longer timespan.
\end{remark}

\begin{remark}\label{remark:shuffle}
For a robot team with $n$ robots, there will be $2^n$ possible combination of hybrid modes of $n$ cables. 
In order to avoid optimizing $2^n$ number of trajectories at the same time, which is not scalable for a large $n$ ($n>3$), we can optimize a \textit{random} subset of the entire hybrid mode combinations during each replanning.
In this way, we \textit{shuffle} the possible hybrid mode combinations among replans.   
\end{remark}

\subsection{Parallelized Optimization for Collaborative Autonomy}
~\label{subsec:centralized_planner}
\begin{algorithm}[t]
\caption{Parallelized Centralized Trajectory Optimization}
\begin{algorithmic}[1]
\label{algorithm:nlp_parallel}
\begin{small}
 \renewcommand{\algorithmicrequire}{\textbf{Input:}}
 \renewcommand{\algorithmicensure}{\textbf{Output:}}
 \REQUIRE Current system state $\mathbf{x}_{\text{init}}$, target state $\mathbf{x}_{\text{goal}}$, obstacles $\mathbb{O}$
 \ENSURE  Optimized state trajectory $\Gamma^{*} = \{\mathbf{x}^*_k\}$, $k=1,2,\dots,N$
 
    \FOR{$\bm{\delta} \in \{0,1\}^{n}$, in parallel,} 
        \STATE 
        $\Gamma(\bm{\delta}) = \{\mathbf{x}\}\leftarrow$ solution of: \\
        $\min_{\mathbf{x}, \mathbf{u}, \mathbf{t}} \  J(\mathbf{x}, \mathbf{u}, \mathbf{t}, \mathbf{x}_{\text{init}}, \mathbf{x}_{\text{goal}})$ \hfill(9a)\label{eq:centralized_cost}\\
        $\quad\textrm{s.t.} 
         \ \ \mathbf{x}_k \in \mathcal{X}_{adm}$ \hfill(9b)\\
        $\qquad\quad \mathbf{u}_k \in \mathcal{U}_{adm} $ \hfill(9c)\\
        $\qquad\quad \mathbf{x}_{k+1} = \mathbf{x}_k + (f_k + f_{k+1})\Delta t/2$ \hfill(9d)\\
        $\qquad\quad \forall i=1\dots n$ \\         
        $\qquad\qquad \textbf{if}\ \bm{\delta}(i)=0,\ T_{i,k}=0, \ (l_0-l_{i,k})\geq 0$ \hfill(9e.0)\\   
        $\qquad\qquad \textbf{if}\ \bm{\delta}(i)=1,\ T_{i,k}>0, \ (l_0-l_{i,k})=0$ \hfill(9e.1)\\
        $\qquad\quad \operatorname{dist}(\mathbb{E}(\mathbf{x}_k), \mathbb{O}) > d$ \hfill(9f)
    \ENDFOR
    \STATE        
    $\Gamma^* = \min_{\bm{\delta}} C(\Gamma(\bm{\delta}))$
 \RETURN $\Gamma^*$ 
\end{small}
\end{algorithmic}
\end{algorithm}
\vspace{-0.5cm}

Having fixed the binary vector $\bm{\delta}$ for the slack/taut mode of each cable, using Algorithm~\ref{algorithm:nlp_parallel}, we can solve the trajectory optimization for the centralized planner to obtain the profile of the states $\mathbf{x}$, input $\mathbf{u}$ and cable tensions $\mathbf{t}$, as defined in Sec.~\ref{sec:model}, of the robots and the cable-towed load.

As illustrated in Algorithm~\ref{algorithm:nlp_parallel}, given the system initial state $\mathbf{x}_{\text{init}}$, target state $\mathbf{x}_{\text{goal}}$, and obstacles $\mathbb{O}$ nearby, and a fixed hybrid mode $\bm{\delta}$, the trajectory optimization in (9) can be solved for a horizon of $N$ indexed by $k$ where each step lasts $\Delta t$ seconds. $\mathcal{X}_{adm}$, $\mathcal{U}_{adm}$ are state and input admissible sets defined and explained in Sec~\ref{subsubsec:state-and-input-constraint}. $T_{i,k}$ and $l_{i,k}$ are Tension $T_i$ and cable length $l_i$ at prediction step $k$.

\subsubsection{Cost Function}

The cost function has the form 
{\small
\begin{equation}\tag{10}
\begin{aligned}
     J &{=}||\mathbf{x}_1{-}\mathbf{x}_{\text{init}} ||^2_{Q_{\text{init}}}{+}||\mathbf{x}_N{-} \mathbf{x}_{\text{goal}} ||^2_{Q_{\text{goal}}}  {+}\sum_{k=1}^{N-1}  (||\mathbf{x}_{k+1}||^2_{Q_\mathbf{x}} \\
      & + ||\mathbf{u}_k||^2_R + ||\mathbf{t}_k||^2_{Q_\mathbf{t}} + \sum_{i=1}^n (l_0-l_{i,k})^2 ),
\end{aligned}
\end{equation}
}where $Q_{\text{init}}$, $Q_{\text{goal}}$, $Q_{\mathbf{x}}$, $R$, $Q_\mathbf{t}$ are positive definite diagonal weight matrices, respectively, and $||\mathbf{x}||^2_Q = \mathbf{x}^T Q \mathbf{x}$. 

The objective function is designed to penalize the state $\mathbf{x}$ and the difference between the robot-load separation $l_i$ and a nominal cable length $l_0$. 
The cable tension $\mathbf{t}$ and virtual input $\mathbf{u}$ are also minimized in order to encourage each robot in the team to contribute to the towing task.
Moreover, we encourage the load to reach the given local target $\mathbf{x}_{\text{goal}}$ at  $\mathbf{x}_N$ which is the end of the trajectory. 
According to Sec.~\ref{sec:framework}, the target $\mathbf{x}_{\text{goal}}$ is from the global path and contains the desired pose for the load. 
Additionally, we introduce a cost to diminish the gap between the first node $\mathbf{x}_1$ to be equal to the current system state $\mathbf{x}_{\text{init}}$. Note that we soften the initial condition constraint by adding it to the cost, so that the planner can still generate different hybrid mode solutions when the robot current state disagrees with the hybrid mode to optimize.

\subsubsection{State and Input Constraint}
\label{subsubsec:state-and-input-constraint}
The states $\mathbf{x}$ and inputs $\mathbf{u}$ should stay within the admissible set $\mathcal{X}_{adm}$ and $\mathcal{U}_{adm}$ via (9b), (9c), respectively.
These constraints not only impose the velocity and acceleration limits to the robots and the towed load, but restrict each robot to move within a nominal region by introducing the boundaries for the robot-load relative orientation $\theta \in [-90^{\circ}, 90^{\circ}]$ and robot-load separation $l \in [l_{min}, l_0]$.

\subsubsection{Dynamics Constraint}
To obtain dynamically feasible trajectories, we impose the system dynamics constraint utilizing trapezodial collocation method via (9d). 
In (9d), $f_k$ represents the system dynamics $f(\mathbf{x}_k, \mathbf{u}_k, \mathbf{t}_k)$ developed in Eqns.~\eqref{eq:model-dyn-pL}\eqref{eq:model-dyn-th}\eqref{eq:model-dyn-p_ri}, while $f_{k+1}$ is $f(\mathbf{x}_{k+1}, \mathbf{u}_{k+1}, \mathbf{t}_{k+1})$.

\subsubsection{Cable Tension and Length Constraint}
Since the hybrid mode represented by the binary vector $\bm{\delta}$ is fixed instead of being a decision variable during optimization, there will be only one constraint selected from (9e.0) and (9e.1) and imposed for the relationship between cable tension and robot-load separation.
For the $i$-th robot, if the hybrid mode is predefined as cable being slack ($\bm{\delta}(i)=0$), the corresponding cable tension $T_i$ should be zero and robot-load separation $l_i$ should be less than the cable length.
Otherwise, the cable should be kept in taut mode with $T_i>0$ and $l_0=l_i$. 

\subsubsection{Collision Avoidance Constraint}
\label{subsubsec:obs-avoid}
We tackle the optimization-based collision avoidance problem using the duality-based formulation developed in~\cite{zhang2020optimization}.
Specifically, we model the system as the union of convex regions: the convex polygon that covers the robot team $\mathbb{E}_{rt}$ and the one that bounds the towed load $\mathbb{E}_l$, and model the $m$-th obstacle as a convex compact set $\mathbb{O}^{(m)}$, as discussed in Sec.~\ref{subsec:convex_polygon}.
The obstacle avoidance constraint via (9e) is a disjoint constraint between each pair of system polygons and all obstacles:
{\small
\begin{equation}\tag{11}
\label{eq:duality-obstacle-avoidance}
\begin{aligned}
\operatorname{dist}(\mathbb{E}_{rt,l}(\mathbf{x}), \mathbb{O}^{(m)}) \geq d ~~
    \Leftrightarrow ~~ \exists \lambda, \nu>0, \epsilon\geq 0:\\
    -g_{rt,l}(\mathbf{x})^{T} \nu+\left(-b^{(m)}\right)^{T} \lambda>d-\epsilon \\
    G_{rt,l}(\mathbf{x})^{T} \nu+ A^{(m)T} \lambda=0, \quad\left\|A^{(m)T} \lambda\right\|<1,
\end{aligned}
\end{equation}
}where $\lambda$, $\nu$ are dual variables, and $d$ is the minimum distance that we want to separate the polygons, and $\epsilon$ is the slack variable, representing the penetration distance of these two sets, to make the problem smoother. 
Therefore, we also need to add a term $K\epsilon^2$ with large positive penalty scalar $K$ in the cost function (9a).

\subsubsection{Optimal Trajectory Selection}
As illustrated in Algorithm~\ref{algorithm:nlp_parallel}, there will be $2^n$ above-mentioned trajectory optimization formulated in (9) with different binary vectors running at the same time, which results in $2^n$ trajectories $\Gamma{(\bm{\delta})}$. 
Once all optimizations are solved, the best trajectory $\Gamma^*$ will be selected by finding the optima based on a metric function $C(\Gamma({\bm{\delta}}))$ which can be written as:
{\small
\begin{equation}\tag{12}
    C(\Gamma({\bm{\delta}})) = ||\mathbf{x}_1(\bm{\delta}) - \mathbf{x}_{\text{init}} ||^2_{Q_{\text{init}}}  + ||\mathbf{x}_N(\bm{\delta}) - \mathbf{x}_{\text{goal}} ||^2_{Q_{\text{goal}}}
\end{equation}
}where we want to find one trajectory $\Gamma^*=\min_{\bm{\delta}} C(\Gamma(\bm{\delta}))$ that is close to current system state while can reach the target state. 
$\Gamma^*$ will therefore be the generated optimal trajectory from the centralized planner. 

\subsection{Decentralized Planner for Each Robot}\label{sec:decentralized_planner}
We now consider the path tracking problem for each robot in the team after receiving the planned state trajectory from the centralized planner.

For each robot, a local goal is found on the centralized trajectory planned for it, and the local goal is updated to be in a short range ahead of the robot, \textit{e.g.}, 0.3~m ahead of the robot. 
For the $i$-th robot, we use trajectory optimization to solve for a horizon of states $\mathbf{x}_{r_i}$ defined in~\eqref{equ:defn_robot_x}. 
In this planner, we want to minimize the difference between the final node and the target goal while satisfying input and state constraints, along with an initial condition constraint for the current system state.
Considering the robot dynamics, we simplify each robot as a double integrator, whose transfer function can be written as $1/s^2$.
The dynamics constraint is imposed by collocation with the trapezoidal method.
In terms of the obstacle avoidance constraint, we keep using the duality-based format like~\eqref{eq:duality-obstacle-avoidance}, but we consider a finer shape of the robot, \textit{i.e.}, a convex bounding box. 
Moreover, the obstacles for the $i$-th ego robot not only includes the obstacles in the environment, but also the load and other robots. 

The outputs of this set of decentralized planners are desired sagittal walking velocity $\dot{x}_{r_i}$, lateral waling velocity $\dot{y}_{r_i}$, and turning yaw rate $\dot{\phi}_{r_i}$, for each robot at current time, and these planners will be updated at a relative high frequency.

\subsection{Optimization Problem Solving}
The above-mentioned optimization is formulated by CasADi with IPOPT as the solver. 
The parallelized  centralized optimization is deployed on a regular personal laptop with a 8-core CPU.  
The decentralized trajectory generation for each robot can be solved on its own computer.

\section{System Integration}
\label{sec:system}
\begin{figure}
    \centering
    \includegraphics[width=0.8\linewidth]{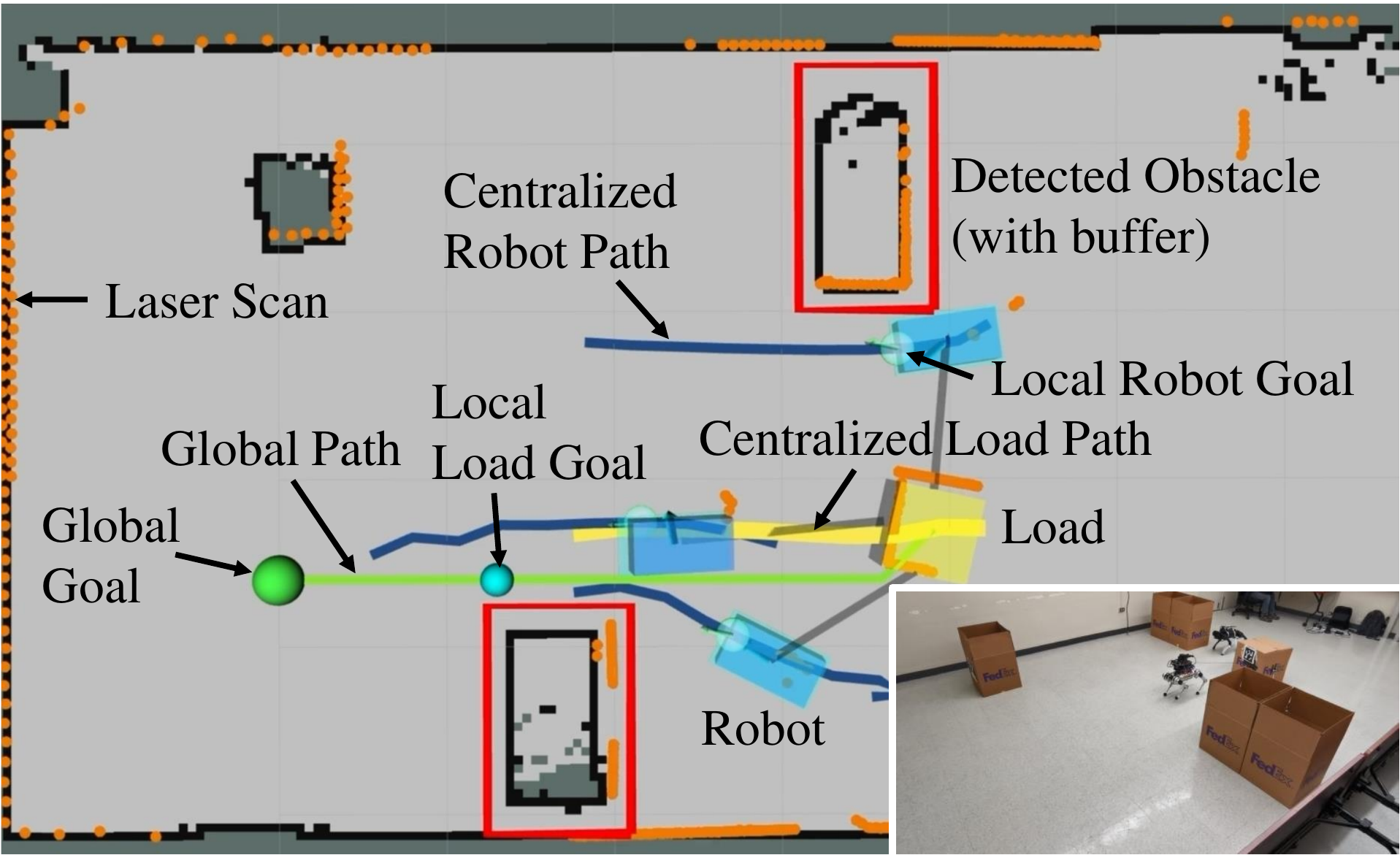}
    \caption{
    Illustration of the environment representation of our system. Global goal for the load is represented by a green point. The global path is drawn as a green line. The local load goal (blue point) always updates to be 2m in front of the load (yellow box) on the global path. Each robot's local goal (blue point) is always kept to be 0.3-meter ahead of its position (blue box) on its path (blue line). The obstacles and laser scans are marked as red rectangles and orange dots.}
    \label{fig:rviz}
    \vspace{-0.3cm}
\end{figure}

As discussed in Sec.~\ref{sec:framework}, the above-introduced optimization-based trajectory planners are embedded in an autonomy that includes a global planner, obstacle detection, state estimator for the load and the robots. 
These are discussed in this section.

\paragraph{Global Planner}~\label{subsec:global_planning}
The global planner utilizes A* search in the three dimensional discretized load configuration space and only considers to find a path for the cable-towed load on an occupancy grid map to reach the given location.

\paragraph{Obstacle Detection} 
The obstacles used in the optimization-based planning are assumed to be convex regions. 
Therefore, we group the neighboring $2D$ laser points into bounding boxes through Euclidean Distance Clustering. 
In this way, the clustered bounding boxes can represent the obstacles in the environment, as shown in Fig.~\ref{fig:rviz}.

\paragraph{Load Detection, Tracking and Following}~\label{subsec:load_following}
As shown in Fig.~\ref{fig:rviz}, we put an AprilTag on one surface of the load, and in this way, the load pose can be represented by the AprilTag pose.
In order to detect the load pose in real-time, we deploy a two DoF gimbal on the back of one robot, and attach a RGB-Depth camera on its end effector which enables yaw and pitch rotations. 
The tag pose is then obtained by robot vision.
The cable-towed load is later tracked by a Kalman Filter, and the gimbal follows the estimated tag position to keep it always inside the camera view.

\paragraph{Robot Localization}~\label{subsec:localization}
There are two methods to estimate the odometry for each robot in this work: one is via VIO by a tracking camera, another one is by $2D$ laser scan matching with a pre-built occupancy map. 
Each robot, depending on the sensor it is equipped with, runs its localization by one of these methods individually.  
Please note that if all of the robots use VIO to localize, the entire collaborative autonomy will not need a pre-built map for the cascaded planning and obstacle detection.

\paragraph{Locomotion Controller}
Each robot is running a model predictive control based locomotion controller~\cite{kim2019highly} to generate motor torques while maintaining trotting gait stability and tracking the velocity commands from decentralized planners. 

In this way, we have developed all the components in the collaborative autonomy proposed in Sec.~\ref{sec:framework}, which includes a global planner, parallelized centralized planner for the team, decentralized planners for each robot, locomotion controllers, obstacle detection, load and robot localization. 
Such a system is validated in the experiments presented in the next section.

\section{Simulation and Experimental Validation}
In this section, we first validate the proposed collaborative autonomy in simulation with different number of robots, then deploy the proposed framework on a robot team constituted by three quadrepedal robots in the real world. The simulation and experimental results are recorded in the video\footnote{\url{https://youtu.be/mBkz_vGOSBs}}.

\subsection{Simulation Validation}
\begin{figure}[]
    \centering
    \includegraphics[width=0.8\linewidth]{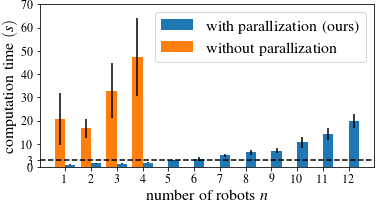}
    \caption{
     Average computation time of replanning using the centralized planner with and without using the proposed parallelization for different robot numbers in the team. Using the proposed paralleization, the average replanning time is below $3$ seconds when $n<6$, which is practical for real-time deployment. In comparison, without the proposed parallelization for hybrid mode switches, the computation time is very high and reaches $30$ seconds with even $3$ robots.
    }
    \label{fig:sim_comp}
    \vspace{-0.4cm}
\end{figure}

We test the scalability of the proposed parallelized planner on multiple robots. We show that when the number of robots is less than six, the proposed parallelized trajectory optimization scheme is able to be solved within 3 seconds.
However, we still remark that the computational time will grow dramatically with the number of robots, as shown in Fig. \ref{fig:sim_comp}, for example it requires 20 seconds to optimize with 12 robots.
Details of this simulation are discussed in Appendix~\ref{appendix-sec:simulation-validation}.

\begin{figure*}[ht]
    \centering
    \begin{subfigure}{0.68\linewidth}
        \centering
        \includegraphics[width=\textwidth]{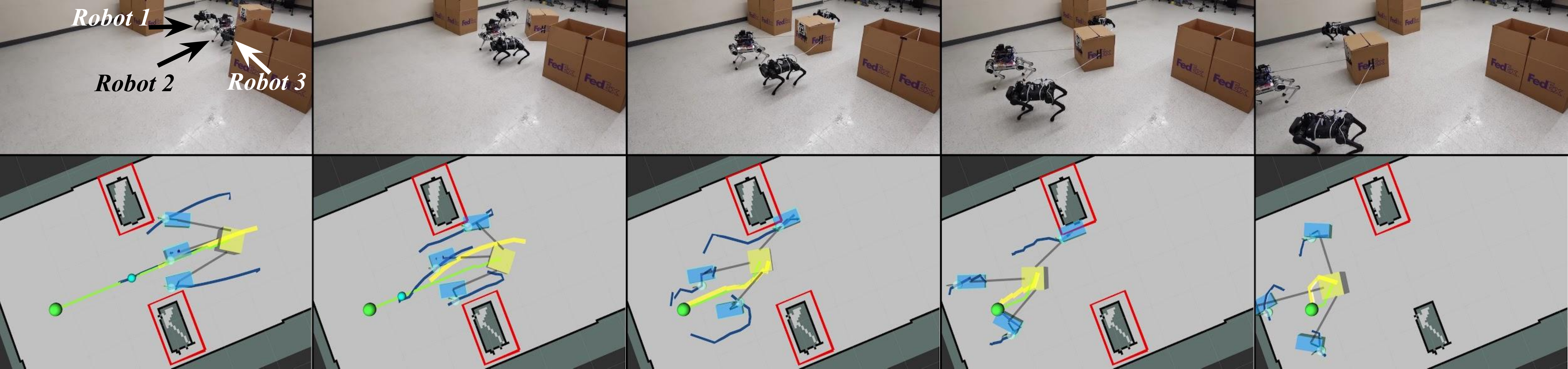}
        \caption{Experiment of collaborative navigation and manipulation through narrow gap with a heavy load.}
        \label{fig:experiment-big-mapA-heavy}
    \end{subfigure}
    \begin{subfigure}{0.22\linewidth}
         \centering
         \includegraphics[width=\textwidth]{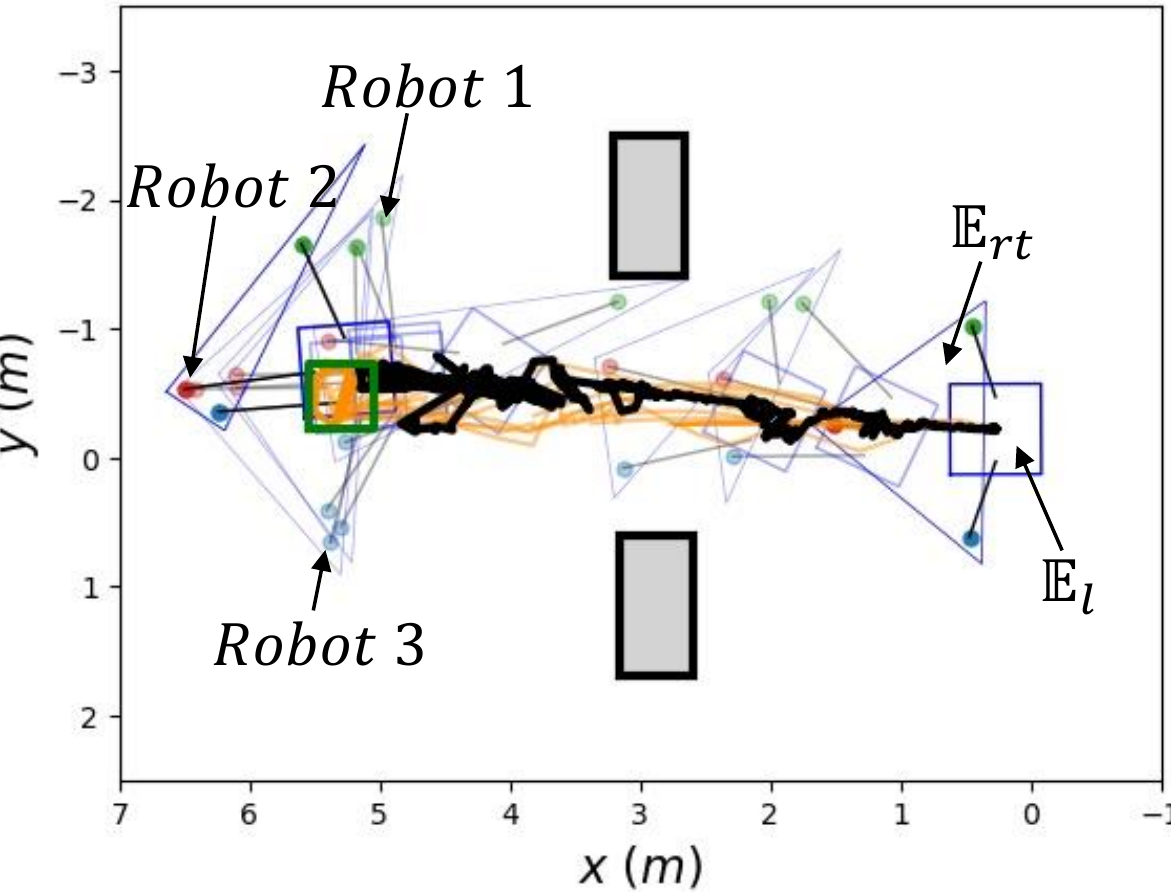}
         \caption{Recorded path of team in (c).}        
         \label{fig:experiment-mape-centralplan-heavy}
    \end{subfigure}
     \begin{subfigure}{0.68\linewidth}
         \centering
         \includegraphics[width=\textwidth]{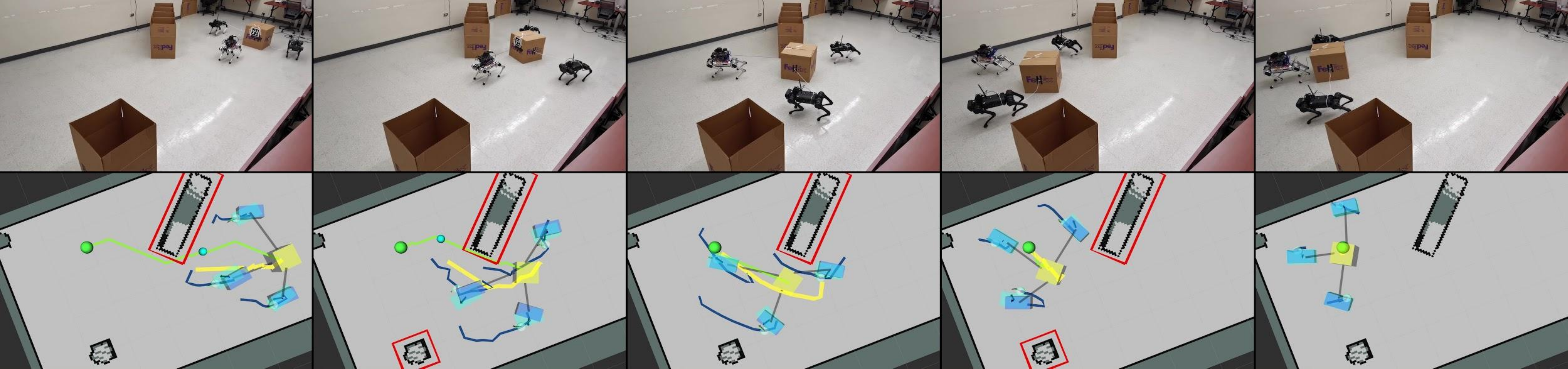}
         \caption{Experiment of collaborative navigation and manipulation through a diagonal gap.}
         \label{fig:experiment-big-mapC}
     \end{subfigure}
     \begin{subfigure}{0.22\linewidth}
         \centering
         \includegraphics[width=\textwidth]{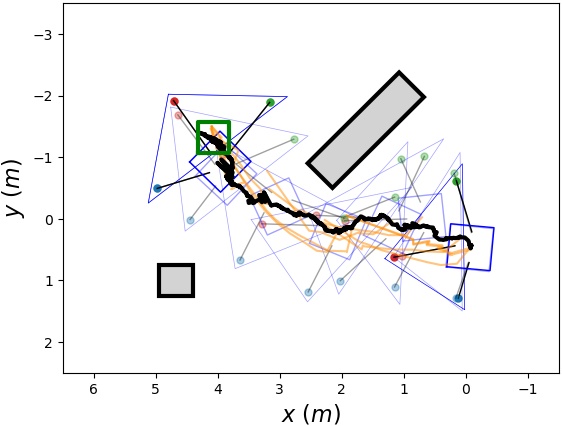}
         \caption{Recorded path of team in (c).}
         \label{fig:experiment-mapc-centralplan}
     \end{subfigure}
    \begin{subfigure}{0.68\linewidth}
        \centering
        \includegraphics[width=\textwidth]{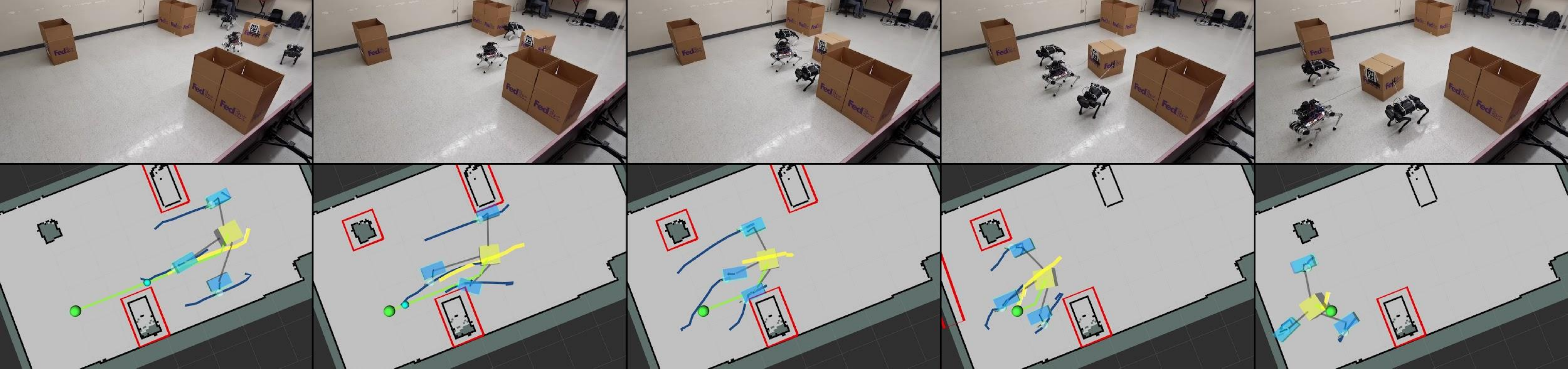}
        \caption{Experiment of collaborative navigation and manipulation through a cluttered space.}
        \label{fig:experiment-big-mapA}
    \end{subfigure}
    \begin{subfigure}{0.22\linewidth}
         \centering
         \includegraphics[width=\textwidth]{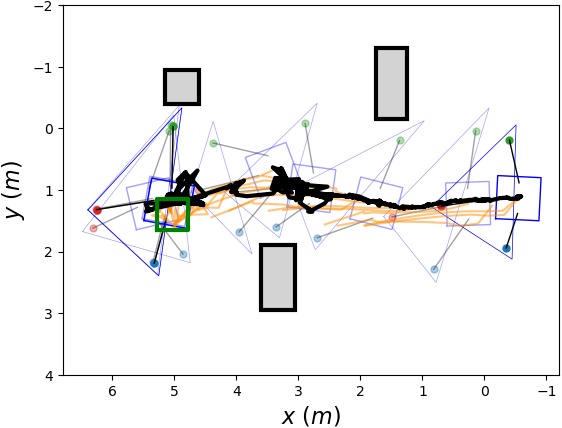}
         \caption{Recorded path of team in (e).}        
         \label{fig:experiment-mapa-centralplan}
    \end{subfigure}

    \caption{Experimental validation of our system in three scenarios, through a narrow gap, a diagonal gap and a cluttered space with the snapshots of the experiments according to the internal environment representation in Fig~\ref{fig:rviz} on the left. 
    The right column figures showcase the recorded plan and actual path of the collaborative team. 
    The centralized planned path for the load is marked as light orange, and the black path is the actual estimated load path.
    The estimated robot positions are also recorded as points. The corresponding convex region $\mathbb{E}_{rt}$ for the robot team and convex load region $\mathbb{E}_l$ are drawn as blue polygons. 
    The detected obstacles are grey bounding boxes.
    During the experiments, the autonomy manifests the ability to avoid collision between the obstacles for both the load and robots themselves while collaborating to tow the load to the goal. 
    The autonomy is also able to switch the cable between slack and taut to traverse narrow spaces.  
    }\label{fig:experiments}
    \vspace{-0.3cm}
\end{figure*}

\subsection{Experimental Setup}
We deploy the proposed collaborative autonomy on three quadrupedal robots, a Mini Cheetah~\cite{katz2019mini}, and two A1s.
The load is a $0.5\times 0.5\times 0.5$~$m^3$ box.
The entire team, including robots and the load, has a width of $3$~m and length of $1.6$~m, when all cables are taut, as demonstrated in the video attached.
Each robot is equipped with a $2D$ LiDAR and one onboard computer running the decentralized planner and obstacle detection.
The Mini Cheetah also has a tracking camera, Intel T265 for VIO, and carries the $2$ DoF gimbal with Intel D435i RGB-Depth camera for load following.  
The global planner and parallelized centralized planner runs on a remote computer.

For convenience, we name the robots anti-clockwise: the robot (A1) to the right of the load is \textit{Robot 1}, the Mini Cheetah that is in front of the load is \textit{Robot 2}, the robot (A1) to the left of the load is \textit{Robot 3}.
The proposed collaborative autonomy is required to manipulate a load and we uses two kinds of loads in experiments: a nominal load ($1$ kg) which a single robot can barely move, and a heavy load ($5$ kg, which is half of each robot's weight) that a single robot is unable to move. We validate the proposed method in three different environments: a cluttered space, a narrow gap formed by two walls, and a diagonal gap formed by a slanted wall, as shown in Fig.~\ref{fig:experiments}.

\subsection{Experimental Validation}
\subsubsection{Navigation and Manipulation through a Narrow Gap}~\label{experiment-big-mapA-heavy}
This scenario entails a row of obstacles with a $2.2$ meter gap, which is narrow compared to the size of the robot team with all taut cables, in the middle of the map as shown in Fig.~\ref{fig:experiment-big-mapA-heavy}.
The collaborative autonomy needs to tow the load through the gap to reach the goal. The load is of 5 kg and is too heavy to move using a single robot. 
As recorded in Fig.~\ref{fig:experiment-mape-centralplan-heavy}, all of the robots engage in taut mode to tow the load before the doorway in order to distribute cable tension and for faster movements. The team then begins to rotate, \textit{Robot 1} decelerates and switches the cable to slack mode to avoid obstacles.
After \textit{Robot 1} circumvents the obstacles, the robots tow the load collaboratively to the goal.
This experiment highlights the necessity of using multiple robots as a team to move a heavy load by the proposed method.
An experiment with the nominal load is also recorded in Fig.~\ref{fig:experiments_narrow_gap}.

\subsubsection{Navigation and Manipulation through a Diagonal Gap}
This scenario consists of a diagonal gap formed by a slanted wall with an obstacle, as showcased in Fig.~\ref{fig:experiment-big-mapC}.
The goal position is behind the slanted wall next to the obstacle, and the load is of 1 kg (nominal).
As presented in Fig.~\ref{fig:experiment-mapc-centralplan}, the robots first collaborate to pull the load until \textit{Robot 1} reaches the wall.
The entire autonomy then rotates the load and itself clockwise to avoid collision with the wall.
During the rotation, \textit{Robot 1} switches the cable to slack mode to prevent dragging the load to the wall and bypasses the wall.
Afterwards, the team rotates back and decelerates towards the goal location.

\subsubsection{Navigation and Manipulation in Cluttered Space}
In this scenario, there are three obstacles distributed in the space, as presented in Fig.~\ref{fig:experiment-big-mapA}.
The collaborative team is required to tow the nominal load through these three obstacles and reach the goal that is past the last obstacle.
After obtaining the goal location, as shown in the Fig.~\ref{fig:experiment-big-mapA}, \textit{Robot 1} first engages the cable slack mode and moves left in order to avoid the first obstacle while the other two robots utilize taut cables to pull the load to minimize effort.
Next, \textit{Robot 3} slows down and makes its cable slack in order to avoid the second obstacle, and \textit{Robot 1} starts to catch up with the team.
At this moment, only \textit{Robot 2} is contributing to the task.
After the team passes the second obstacle, \textit{Robot 2} starts to slow down and makes its cable slack while the other two robots pull to rotate the load and move left to reach the goal.

In all of the above-mentioned experiments, using the proposed method, the robotic team demonstrates a considerable level of autonomy of collaboration to manipulate the load via cables to the goal.
The proposed collaborative autonomy showcases the advantages of utilizing multiple robots and hybrid physical interaction.
The robots are able to work together and contribute to the towing action with all cables taut when in free space, and cables going slack during re-positioning and re-orienting to avoid collisions without affecting load movement.
In this way, the robots can traverse through narrow spaces that are otherwise difficult to pass if the cables are constantly taut.

\section{Conclusion and Future Work}
In this work, we develop a navigation autonomy that combines collaborative manipulation and hybrid physical interaction to move and navigate a cable-towed load in cluttered environments with multiple legged robots.
The parallelization for optimization with different hybrid modes and the decentralized optimization for path tracking of each single robot are applied, and multiple robots can be used to navigate and manipulate a cable-towed load through various cluttered environments.

However, the proposed centralized method suffers from the curse of dimensionality due to the increasing complexity of the constrained optimization with the growth of the number of robots.
In the future, developing an end-to-end decentralized planner that runs on-board each agent in such robotic teams with cable-towed load could be an interesting research direction to increase the usability and agility of such collaborative robotic teams with quadrupedal robots.

\section*{Acknowledgements} 
This work was partially supported through a BAIR-FAIR collaboration, Hong Kong Center for Logistics Robotics, and NSF Grant CMMI-1931853. We thank Prof. Francesco Borrelli for insightful suggestions, and also thank Prof. Sangbae Kim, the MIT Biomimetic Robotics Lab, and NAVER LABS for lending the Mini Cheetah for experiments.

\bibliographystyle{IEEEtran}
\bibliography{bib/bibliography}{}

\begin{thebibliography}{10}
\providecommand{\url}[1]{#1}
\csname url@samestyle\endcsname
\providecommand{\newblock}{\relax}
\providecommand{\bibinfo}[2]{#2}
\providecommand{\BIBentrySTDinterwordspacing}{\spaceskip=0pt\relax}
\providecommand{\BIBentryALTinterwordstretchfactor}{4}
\providecommand{\BIBentryALTinterwordspacing}{\spaceskip=\fontdimen2\font plus
\BIBentryALTinterwordstretchfactor\fontdimen3\font minus
  \fontdimen4\font\relax}
\providecommand{\BIBforeignlanguage}[2]{{%
\expandafter\ifx\csname l@#1\endcsname\relax
\typeout{** WARNING: IEEEtran.bst: No hyphenation pattern has been}%
\typeout{** loaded for the language `#1'. Using the pattern for}%
\typeout{** the default language instead.}%
\else
\language=\csname l@#1\endcsname
\fi
#2}}
\providecommand{\BIBdecl}{\relax}
\BIBdecl

\bibitem{katz2019mini}
B.~Katz, J.~Di~Carlo, and S.~Kim, ``Mini cheetah: A platform for pushing the
  limits of dynamic quadruped control,'' in \emph{Int. Conf. on Robotics and
  Automation}, 2019, pp. 6295--6301.

\bibitem{kim2019highly}
D.~Kim, J.~Di~Carlo, B.~Katz, G.~Bledt, and S.~Kim, ``Highly dynamic quadruped
  locomotion via whole-body impulse control and model predictive control,''
  \emph{arXiv preprint arXiv:1909.06586}, 2019.

\bibitem{yang2020dynamic}
C.~Yang, B.~Zhang, J.~Zeng, A.~Agrawal, and K.~Sreenath, ``Dynamic legged
  manipulation of a ball through multi-contact optimization,'' in \emph{Int.
  Conf. on Intelligent Robots and Systems}, 2020, pp. 7513--7520.

\bibitem{gilroy2021autonomous}
S.~Gilroy, D.~Lau, L.~Yang, E.~Izaguirre, K.~Biermayer, A.~Xiao, M.~Sun,
  A.~Agrawal, J.~Zeng, Z.~Li \emph{et~al.}, ``Autonomous navigation for
  quadrupedal robots with optimized jumping through constrained obstacles,'' in
  \emph{Int. Conf. on Automation Science and Engineering}, 2021.

\bibitem{kanazawa2019adaptive}
A.~Kanazawa, J.~Kinugawa, and K.~Kosuge, ``Adaptive motion planning for a
  collaborative robot based on prediction uncertainty to enhance human safety
  and work efficiency,'' \emph{Transactions on Robotics}, 2019.

\bibitem{xiao2021robotic}
A.~Xiao, W.~Tong, L.~Yang, J.~Zeng, Z.~Li, and K.~Sreenath, ``Robotic guide
  dog: Leading a human with leash-guided hybrid physical interaction,'' in
  \emph{Int. Conf. on Robotics and Automation}, 2021.

\bibitem{liniger2019noncooperative}
A.~Liniger and J.~Lygeros, ``A noncooperative game approach to autonomous
  racing,'' \emph{Transactions on Control Systems Technology}, 2019.

\bibitem{cleac2021algames}
L.~Cleac’h, M.~Schwager, Z.~Manchester \emph{et~al.}, ``Algames: a fast
  augmented lagrangian solver for constrained dynamic games,'' \emph{Autonomous
  Robots}, pp. 1--15, 2021.

\bibitem{tung2021learning}
A.~Tung, J.~Wong, A.~Mandlekar, R.~Mart{\'\i}n-Mart{\'\i}n, Y.~Zhu, L.~Fei-Fei,
  and S.~Savarese, ``Learning multi-arm manipulation through collaborative
  teleoperation,'' in \emph{Int. Conf. on Robotics and Automation}, 2021.

\bibitem{sanchez2021four}
D.~Sanchez, W.~Wan, and K.~Harada, ``Four-arm collaboration: Two dual-arm
  robots work together to manipulate tethered tools,'' \emph{Transactions on
  Mechatronics}, 2021.

\bibitem{wang2018cooperative}
Z.~Wang, S.~Singh, M.~Pavone, and M.~Schwager, ``Cooperative object transport
  in 3d with multiple quadrotors using no peer communication,'' in \emph{Int.
  Conf. on Robotics and Automation}, 2018, pp. 1064--1071.

\bibitem{sanalitro2020full}
D.~Sanalitro, H.~J. Savino, M.~Tognon, J.~Cort{\'e}s, and A.~Franchi,
  ``Full-pose manipulation control of a cable-suspended load with multiple uavs
  under uncertainties,'' \emph{Robotics and Automation Letters}, 2020.

\bibitem{cheng2009cooperative}
P.~Cheng, J.~Fink, V.~Kumar, and J.-S. Pang, ``{Cooperative Towing With
  Multiple Robots},'' \emph{Journal of Mechanisms and Robotics}, 2008.

\bibitem{wang2016kinematic}
Z.~Wang and M.~Schwager, ``Kinematic multi-robot manipulation with no
  communication using force feedback,'' in \emph{Int. Conf. on Robotics and
  Automation}, 2016, pp. 427--432.

\bibitem{losey2020learning}
D.~P. Losey, M.~Li, J.~Bohg, and D.~Sadigh, ``Learning from my partner’s
  actions: Roles in decentralized robot teams,'' in \emph{Conf. on robot
  learning}.\hskip 1em plus 0.5em minus 0.4em\relax PMLR, 2020, pp. 752--765.

\bibitem{loianno2017cooperative}
G.~Loianno and V.~Kumar, ``Cooperative transportation using small quadrotors
  using monocular vision and inertial sensing,'' \emph{Robotics and Automation
  Letters}, vol.~3, no.~2, pp. 680--687, 2017.

\bibitem{culbertson2021decentralized}
P.~Culbertson, J.-J.~E. Slotine, and M.~Schwager, ``Decentralized adaptive
  control for collaborative manipulation of rigid bodies,'' 2021.

\bibitem{wang2017prestressed}
Z.~Wang, Y.~Torigoe, and S.~Hirai, ``A prestressed soft gripper: design,
  modeling, fabrication, and tests for food handling,'' \emph{Robotics and
  Automation Letters}, vol.~2, no.~4, pp. 1909--1916, 2017.

\bibitem{herguedas2019survey}
R.~Herguedas, G.~L{\'o}pez-Nicol{\'a}s, R.~Arag{\"u}{\'e}s, and
  C.~Sag{\"u}{\'e}s, ``Survey on multi-robot manipulation of deformable
  objects,'' in \emph{Int. Conf. on Emerging Technologies and Factory
  Automation}, 2019.

\bibitem{kim2018bound}
Y.-H. Kim and D.~A. Shell, ``Bound to help: cooperative manipulation of objects
  via compliant, unactuated tails,'' \emph{Autonomous Robots}, 2018.

\bibitem{cardona2021non}
G.~A. Cardona, D.~S. D’Antonio, C.-I. Vasile, and D.~Salda{\~n}a,
  ``Non-prehensile manipulation of cuboid objects using a catenary robot,'' in
  \emph{Int. Conf. on Intelligent Robots and Systems}, 2021.

\bibitem{bhattacharya2015topological}
S.~Bhattacharya, S.~Kim, H.~Heidarsson, G.~S. Sukhatme, and V.~Kumar, ``A
  topological approach to using cables to separate and manipulate sets of
  objects,'' \emph{The Int. Journal of Robotics Research}, 2015.

\bibitem{zeng2019geometric}
J.~Zeng, P.~Kotaru, and K.~Sreenath, ``Geometric control and differential
  flatness of a quadrotor uav with load suspended from a pulley,'' in
  \emph{American Control Conf.}, 2019, pp. 2420--2427.

\bibitem{tognon2018aerial}
M.~Tognon, C.~Gabellieri, L.~Pallottino, and A.~Franchi, ``Aerial
  co-manipulation with cables: The role of internal force for equilibria,
  stability, and passivity,'' \emph{Robotics and Automation Letters}, 2018.

\bibitem{lee2013geometric}
T.~Lee, K.~Sreenath, and V.~Kumar, ``Geometric control of cooperating multiple
  quadrotor uavs with a suspended payload,'' in \emph{Conf. on Decision and
  Control}, 2013, pp. 5510--5515.

\bibitem{wu2014geometric}
G.~Wu and K.~Sreenath, ``Geometric control of multiple quadrotors transporting
  a rigid-body load,'' in \emph{Conf. on Decision and Control}, 2014.

\bibitem{Kotaru2020multiple}
P.~Kotaru and K.~Sreenath, ``Multiple quadrotors carrying a flexible hose:
  dynamics, differential flatness and control,'' \emph{IFAC-PapersOnLine}, pp.
  8832--8839, 2020, 21st IFAC World Congress.

\bibitem{minniti2021adaptive}
M.~V. Minniti, R.~Grandia, F.~Farshidian, and M.~Hutter, ``Adaptive clf-mpc
  with application to quadrupedal robots,'' \emph{Robotics and Automation
  Letters}, 2021.

\bibitem{tang2015mixed}
S.~Tang and V.~Kumar, ``Mixed integer quadratic program trajectory generation
  for a quadrotor with a cable-suspended payload,'' in \emph{Int. Conf. on
  robotics and automation}, 2015, pp. 2216--2222.

\bibitem{zeng2020differential}
J.~Zeng, P.~Kotaru, M.~W. Mueller, and K.~Sreenath, ``Differential flatness
  based path planning with direct collocation on hybrid modes for a quadrotor
  with a cable-suspended payload,'' \emph{Robotics and Automation Letters},
  vol.~5, no.~2, pp. 3074--3081, 2020.

\bibitem{zhang2020optimization}
X.~Zhang, A.~Liniger, and F.~Borrelli, ``Optimization-based collision
  avoidance,'' \emph{Transactions on Control Systems Technology}, 2020.

\bibitem{he2022autonomous}
S.~He, J.~Zeng, and K.~Sreenath, ``Autonomous racing with multiple vehicles
  using a parallelized optimization with safety guarantee using control barrier
  functions,'' in \emph{Int. Conf. on Robotics and Automation}, 2022.

\end{thebibliography}

\appendix
\subsection{Simulation Validation}
\label{appendix-sec:simulation-validation}

\begin{figure}[h]
\centering
\begin{subfigure}{0.32\linewidth}
  \centering
  \includegraphics[width=\linewidth]{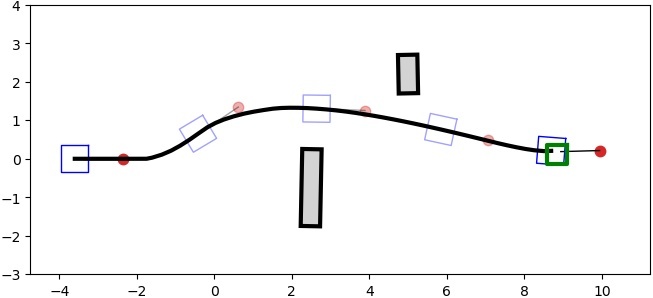}
  \caption{1-robot team}
  \label{subfig:1_robot}
\end{subfigure}
\begin{subfigure}{0.32\linewidth}
  \centering
  \includegraphics[width=\linewidth]{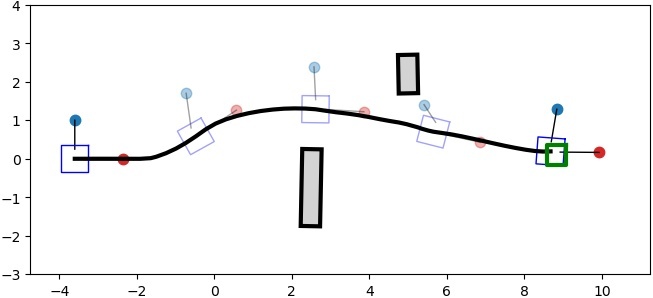}
  \caption{2-robot team}
  \label{subfig:2_robot}
\end{subfigure}
\begin{subfigure}{0.32\linewidth}
  \centering
  \includegraphics[width=\linewidth]{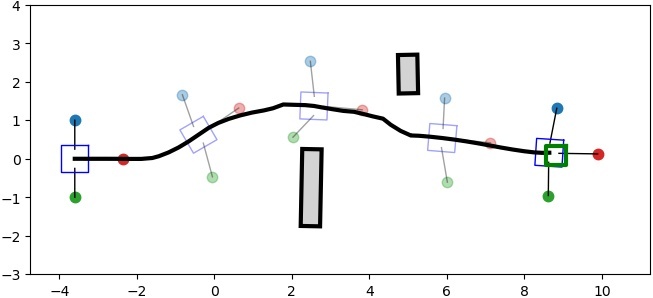}
  \caption{3-robot team}
  \label{subfig:3_robot}
\end{subfigure} \\
\begin{subfigure}{0.32\linewidth}
  \centering
  \includegraphics[width=\linewidth]{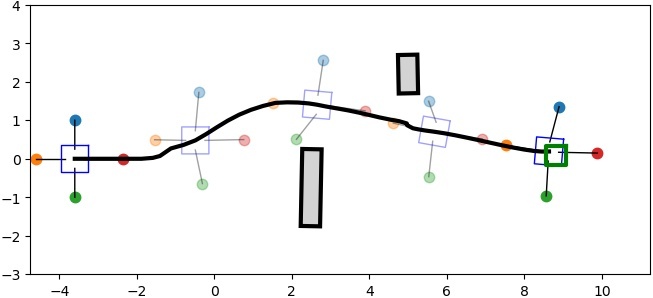}
  \caption{4-robot team}
  \label{subfig:4_robot}
\end{subfigure}
\begin{subfigure}{0.32\linewidth}
  \centering
  \includegraphics[width=\linewidth]{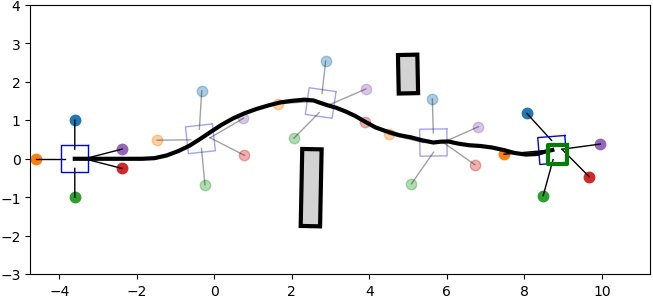}
  \caption{5-robot team}
  \label{subfig:5_robot}
\end{subfigure}
\begin{subfigure}{0.32\linewidth}
  \centering
  \includegraphics[width=\linewidth]{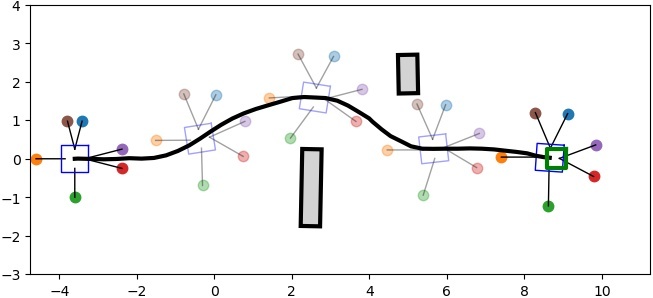}
  \caption{6-robot team}
  \label{subfig:6_robot}
\end{subfigure} \\
\begin{subfigure}{0.32\linewidth}
  \centering
  \includegraphics[width=\linewidth]{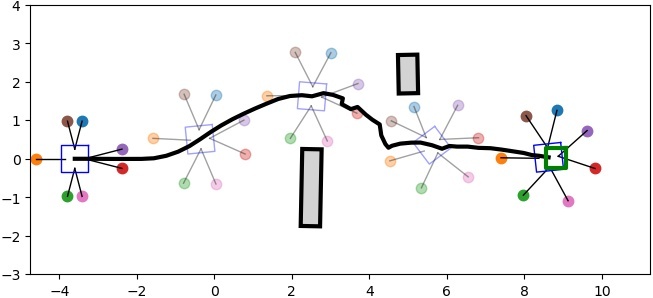}
  \caption{7-robot team}
  \label{subfig:7_robot}
\end{subfigure}
\begin{subfigure}{0.32\linewidth}
  \centering
  \includegraphics[width=\linewidth]{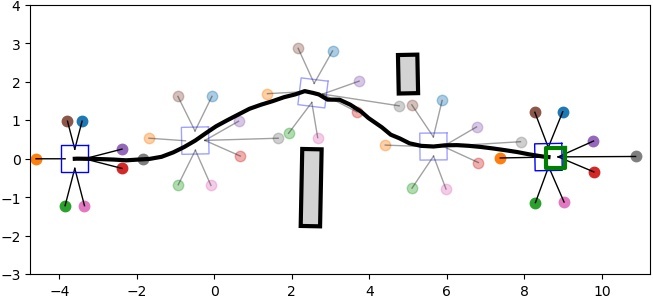}
  \caption{8-robot team}
  \label{subfig:8_robot}
\end{subfigure}
\begin{subfigure}{0.32\linewidth}
  \centering
  \includegraphics[width=\linewidth]{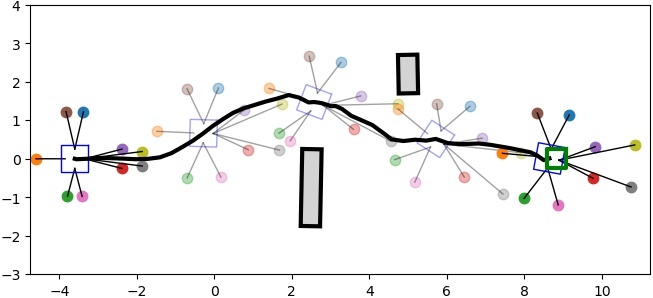}
  \caption{9-robot team}
  \label{subfig:9_robot}
\end{subfigure} \\
\begin{subfigure}{0.32\linewidth}
  \centering
  \includegraphics[width=\linewidth]{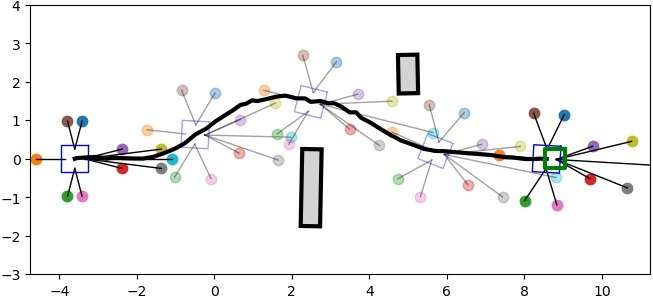}
  \caption{10-robot team}
  \label{subfig:10_robot}
\end{subfigure}
\begin{subfigure}{0.32\linewidth}
  \centering
  \includegraphics[width=\linewidth]{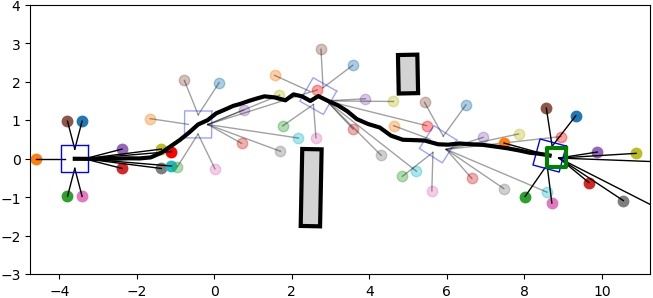}
  \caption{11-robot team}
  \label{subfig:11_robot}
\end{subfigure}
\begin{subfigure}{0.32\linewidth}
  \centering
  \includegraphics[width=\linewidth]{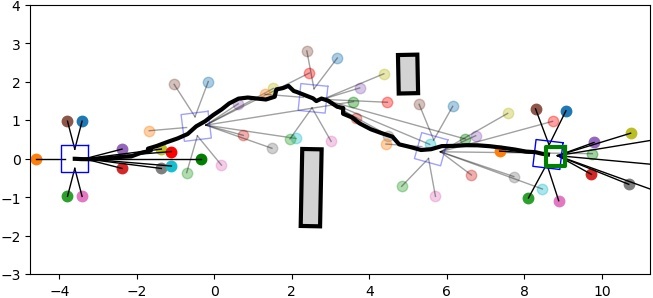}
  \caption{12-robot team}
  \label{subfig:12_robot}
\end{subfigure}
\caption{
The simulated paths of the robot teams with different number of robots travelling through a congested space using the proposed autonomy.
The proposed parallelized centralized planner is able to find collision free path by reactive replanning to enable the robot teams consisted by different number of robots to move the load to the given target which is marked as a green box.
The obstacles are marked as grey regions and the positions of robots are marked as circles with different colors. The recorded history paths of the load are black lines.
}
\label{fig:sim_multi_robot}
\end{figure}

In order to test the scalability of the proposed parallelized centralized planner on multiple robots, the proposed collaborative autonomy is tested in a simulation with the robot teams that have different number (1 to 12) of robots, as shown in Fig.~\ref{fig:sim_multi_robot}. 
In this simulation, the system dynamics is calculated by Eqns.~\eqref{eq:model-dyn-pL},\eqref{eq:model-dyn-th},\eqref{eq:model-dyn-p_ri}. 
During the simulation, the robot team is required to move the load to the goal location (green marker in Fig.~\ref{fig:sim_multi_robot}) while avoiding obstacles. 
A global collision-free path to the goal is firstly obtained, and centralized planner replans online to generate desired trajectory for each robot. 
We made an assumption that each robot can perfectly track the given planned path, which allows us to skip the decentralized planner and locomotion controller for each quadrupedal robot, and to make the simulation computationally feasible on a personal laptop.
We further assume the system state is well known.

Using the proposed method illustrated in Sec.~\ref{subsec:centralized_planner}, the collaborative robot teams with the number of robots varying from 1 to 12 are all able to tow the load to the given location while taking advantage of taut/slack switching of cables to travel through a narrow gap formed by two obstacles, as shown in Fig.~\ref{fig:sim_multi_robot}.
Moreover, the computation time of each replanning against the number of the robots using the parallelized centralized planner during the travel is recorded in Fig.~\ref{fig:sim_comp}. 
For the robot team that has less than 6 robots, the proposed parallelized trajectory optimization scheme is able to solve within $3$ seconds on a single personal laptop.
This makes the algorithm practical to deploy on the robot running in real time. 
If we do not utilize the proposed method and consider hybrid mode switches during each replanning, the average replanning time can be more than $30$ second for just three robots in the same scenario, shown as the orange bars in Fig.~\ref{fig:sim_comp}.
Therefore, the proposed method shows a clear advantage in terms of the computation efficiency. 
However, with the increasing number of robots, the computing time grows dramatically and reaches $20$~second to finish one replanning for the case with 12 robots.
This is due to the increasing complexity of the constrained optimization: there are more robot team polygons needed to be considered for obstacle avoidance and the system dynamics are in higher dimensions. 
We remark that planning for large number ($n\geq6$) of robots is a limitation of the proposed centralized method due to the increased computation time.

\subsection{Experiment Validation}
\begin{figure}[H]
    \centering
    \begin{subfigure}{1.0\linewidth}
         \centering
         \includegraphics[width=\textwidth]{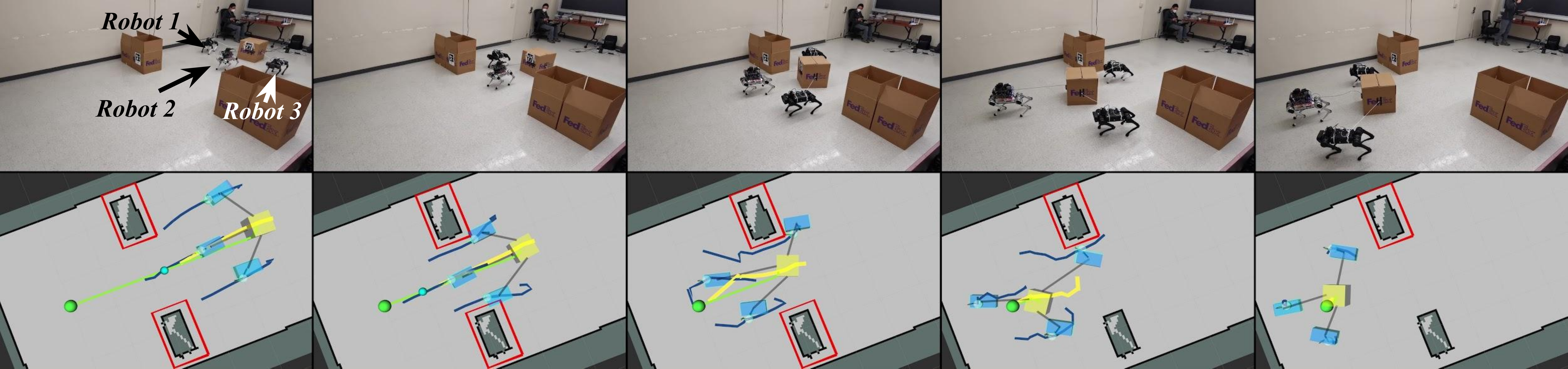}
         \caption{Experimental run of collaborative navigation and manipulation through a narrow gap with a nominal load.}
         \label{fig:experiment-big-mapE}
     \end{subfigure} \\
     \begin{subfigure}{0.5\linewidth}
         \centering
         \includegraphics[width=\textwidth]{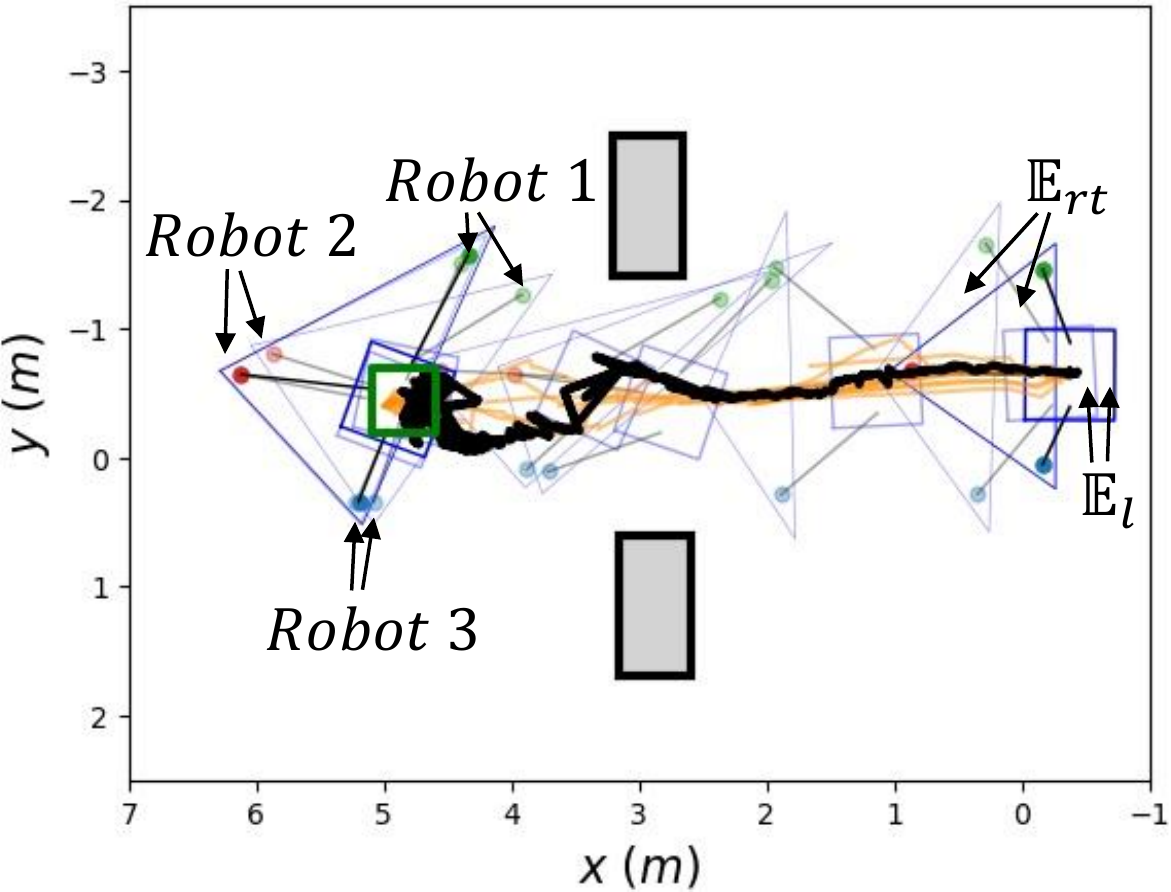}
         \caption{Recorded path of team in (a).}
         \label{fig:experiment-mape-centralplan}
     \end{subfigure}
    \caption{Experiment validation of the proposed system in the narrow gap scenario while manipulating a nominal load of $1$ kg.
    }\label{fig:experiments_narrow_gap}
\end{figure}

The experimental validation for navigation task through a narrow gap with a nominal load ($1$ kg), shown in Fig. \ref{fig:experiments_narrow_gap}.

\end{document}